# Topical Keyphrase Extraction with Hierarchical Semantic Networks[1]

Yoo yeon Sung

yooyeon6@gmail.com

Korea University

Seoung Bum Kim*

sbkim1@korea.ac.kr

Korea University

**Abstract**

Topical keyphrase extraction is used to summarize large collections of text documents. However, traditional methods cannot properly reflect the intrinsic semantics and relationships of keyphrases because they rely on a simple term-frequency-based process. Consequently, these methods are not effective in obtaining significant contextual knowledge. To resolve this, we propose a topical keyphrase extraction method based on a hierarchical semantic network and multiple centrality network measures that together reflect the hierarchical semantics of keyphrases. We conduct experiments on real data to examine the practicality of the proposed method and to compare its performance with that of existing topical keyphrase extraction methods. The results confirm that the proposed method outperforms state-of-the-art topical keyphrase extraction methods in terms of the representativeness of the selected keyphrases for each topic. The proposed method can effectively reflect intrinsic keyphrase semantics and interrelationships.

**Keywords**: topical keyphrase extraction, semantic relationships, hierarchical networks, phrase rankings, text mining

## 1. Introduction

Because of the rapid increase in the number of digital documents, systemized compression of big unstructured text data has become inevitable for efficient retrieval of useful information [1, 2]. Organized knowledge of text data patterns can assist those who lack an overall idea of the domain. Many academic organizations and mass media provide category information that broadly explains the content of documents. However, this information lacks directly associated keyphrases that provide a clear guideline, and obtaining them is a major challenge [3]. Automatic keyphrase extraction methods are expected to provide guidelines for extracting a set of phrases that preserve the main theme of a document [4]. Selecting the appropriate keyphrases is essential in domains that require an effective and fast understanding of large text collections [5].

However, document data tend to involve a large number of distinct topics. Therefore, without adequate topic information, it is difficult to extract keyphrases that represent the entire context. Thus, topical keyphrase extraction methods were developed, where the topic is considered in keyphrase extraction. With topical keyphrase extraction methods, searchers encounter many groups of keyphrases in which each group symbolizes a single topic in the document. In each group, the keyphrases represent the group topic. These are called topical keyphrases and are expected to provide a better overview of the entire document data. Numerous topical keyphrase extraction methods have been developed; they use clustering techniques [6], language modeling [4], statistical modeling, and network-based approaches. Statistical models select topical keyphrases through the inferred topic-word probability or a specific term-frequency rule. These models are primarily intended for statistical interpretation of term specificity [1,7]. However, they cannot capture the underlying meaning of keyphrases because they are primarily based on the computation of

---

[1] Abbreviations:

HSN: Hierarchical semantic network, LDA: Latent Dirichlet allocation, TF–ITF: Term frequency–inverse topic frequency, TPR: Topical PageRank, TR: TextRank, WPR: Weighted PageRank the frequency of keyphrases.

Network-based topical keyphrase extraction methods attempt to determine the semantic relationships of keyphrases. These methods initially construct a phrase network and rank its phrases according to their salience. A phrase network is composed of nodes and edges. Nodes correspond to phrases, and edges represent the relatedness of the nodes they connect. To compute node relatedness, the co-occurrence frequency of nearby words is typically used. For phrase-based TR, a node is directed to the another node when the corresponding word is positioned before other words in a word sequence [8]. For TPR, two nodes are linked if their corresponding words co-occur in a W-sized window. The edge directions are determined by the first word pointing to every other word in the same window, and the edge weights are simply set to the co-occurrence counts within a window [9]. In both methods, PageRank, a network centrality measure, determines the representativeness of a keyphrase [9].

Although phrase evaluation in previous network-based ranking approaches can recognize elementary keyphrase relationships, hierarchical relationships cannot be identified. To address this limitation, we propose an HSN to extract topical keyphrases through centrality measures. Identifying hierarchical keyphrase relationships is motivated by [10], where it is considered important in obtaining the main themes of document data by detecting the phrase in a group that contains the most general information and thus summarizes the group content. For example, the phrase "machine learning" is more general than "supervised learning" in documents concerning machine learning algorithms that include the concept of supervised learning. The hierarchical relationships are obtained by a frequency rule called the "subsumption rule" [10].

In this study, we propose an HSN that selects the most representative topical phrases by considering their complex and hierarchical relationships. Traditional methods cannot explicitly detect such complex and intrinsic relationships, whereas the hierarchical organization in the proposed method enables sorting keyphrases according to their degree of representativeness; thus, we can detect the phrases that best represent each topic within a given set of documents.

The main contributions of this study can be summarized as follows:

(1) The proposed method can reflect the hierarchical relationships between phrases by using an association rule in the construction of a hierarchical semantic network. These relationships identify the edges of the network and determine which phrases should be used as nodes. Only nodes with close hierarchical relationships are added to the network, along with adequate edge directions and edge weights. Unlike other network-based keyphrase extraction methods that use simple co-occurrence rules, the proposed method uses the subsumption rule, which considers the hierarchical relationships between phrases.

(2) We propose topical keyphrase selection by combining the phrase scores by multiple centrality network measures. The representativeness of each node in the phrase network is determined by its inflowing edges and edge weights. Thus, we integrate some centrality measures that can evaluate the nodes according to their inflowing edges. Because these centrality measures reflect the phrase hierarchy, the most highly evaluated phrases can best represent the topics of a document.

(3) To demonstrate the usefulness and applicability of the proposed method, we use real-world large text data for comparison with existing keyphrase extraction methods in terms of effectiveness in selecting representative topical keyphrases. The results confirm that the proposed method outperforms the others.

The remainder of this paper is organized as follows. In Section 2, we review existing keyphrase extraction approaches associated with the present study. In Section 3, we present the details of the proposed HSN. Section 4 presents and discusses the experimental results and their evaluation. Finally, Section 5 concludes the paper.

**2. Related Work**

The use of supervised and unsupervised approaches in topical keyphrase extraction has attracted significant attention. Various supervised learning algorithms, which are often treated as binary classification algorithms, were trained on documents annotated with keyphrases to determine whether a candidate phrase is a keyphrase or not [5]. However, recasting keyphrase extraction as a classification problem suffers from the weakness of treating keyphrases independently of one another in the training process [5, 11]. Consequently, it is difficult to compare them and select the keyphrase that best represents the entire set of documents.

To compensate for the limitations of supervised approaches, unsupervised approaches have been introduced. For example, a key cluster algorithm was proposed to cluster semantically similar candidate phrases using Wikipedia and co-occurrence-based statistics [5, 6]. This method selects the centroid of each topic cluster as a topical keyphrase. However, its drawback is that each topic is given equal importance. In practice, there could be topics that are unimportant and should not have keyphrases representing them [5]. To extract all candidate keyphrases from important topics through Wikipedia, a community cluster algorithm was proposed [5]. However, because the algorithm depends on Wikipedia-based knowledge, it cannot reflect specialized topics in specific document data.

Keyphrase ranking is a different methodology in which candidate phrases are scored and only those with the highest scores are selected as representative topical keyphrases. TF–ITF is a variant of the traditional term frequency–inverse document frequency that is used for statistically ranking the specificity of index terms and document descriptions [3, 7]. Accordingly, the weight of a keyword is the product of the number of occurrences in the document (DF) and the rarity value across the entire document collection (IDF). TF–ITF counts the number of words in the topical document collection instead of a single document. The TF–ITF weights are statistical measures for evaluating the degree to which a word is relevant to the corresponding topic. This implies that highly weighted keyphrases are used as representative keyphrases for the topic. A well-known topic model, LDA, can extract keywords based on a topic model that takes documents as input, models them as mixtures of different topics, and discovers word distributions for each topic [12]. The word probabilities can be interpreted as weights for ranking the degree of relevance of the words to the topics. A word with higher probability in the word distribution is considered to be more representative of the topic, compared with other words in the document collection. Although the keywords and keyphrases resulting from the TF–ITF and LDA algorithms may encompass topic-relevant information, they cannot capture the semantic relationships of keyphrases.

Network-based ranking methods are the state-of-the-art in unsupervised approaches [9, 13]. The TR, TPR, and Node and Edge Rank (NE-Rank) models provide network measures for ranking candidate words [8, 9, 14]. TR generates a word network that considers the co-occurrence of words in the neighboring sequence of a document. Each phrase is then scored by a PageRank network measure [8]. The top-ranked words are post-processed to be combined into phrases if they are adjacent in the text; otherwise, they remain as keywords. TPR constructs a phrase network based on the co-occurrence of phrases in the phrase sequence and scores each phrase using a modified PageRank network measure. TPR includes information on word distribution per topic from LDA and thus yields representative topical keyphrases [9]. The NE-Rank algorithm is a combination of the TR and PageRank network measures with edge weight information. It is designed to extract topical keyphrases from short texts such as tweets on Twitter [14]. The three aforementioned network-based ranking methods have limitations regarding the construction of the phrase network. Because phrase relations in the network are implicated by a simple co-occurrence rule applied to neighboring phrases, hierarchical relationships cannot be comprehended. Moreover, TPR induces a limited selection of keyphrases because it directly adopts the LDA word distribution per topic into the PageRank network measure.

## 3. Proposed Method

### 3.1. Overview of the proposed topical keyphrase extraction method

Network-based topical keyphrase extraction methods consist mainly of four procedures: (1) topical document collection, (2) topical candidate phrase composition, (3) construction of the topical hierarchical semantic network, and (4) evaluation of topical candidate phrases. Steps (3) and (4) are the main contributions of the proposed HSN, which essentially reflects the hierarchical relationships of topical candidate phrases in the construction of the semantic network and phrase evaluation. First, the relationships between topical candidate phrases are identified by a subsumption rule, and an association rule is used to define the nodes and edges of the topical HSN. Then, several centrality network measures are used to evaluate the phrases in the network. Finally, the most highly evaluated topical keyphrases are extracted. Fig. 1 shows an overview of the proposed HSN.

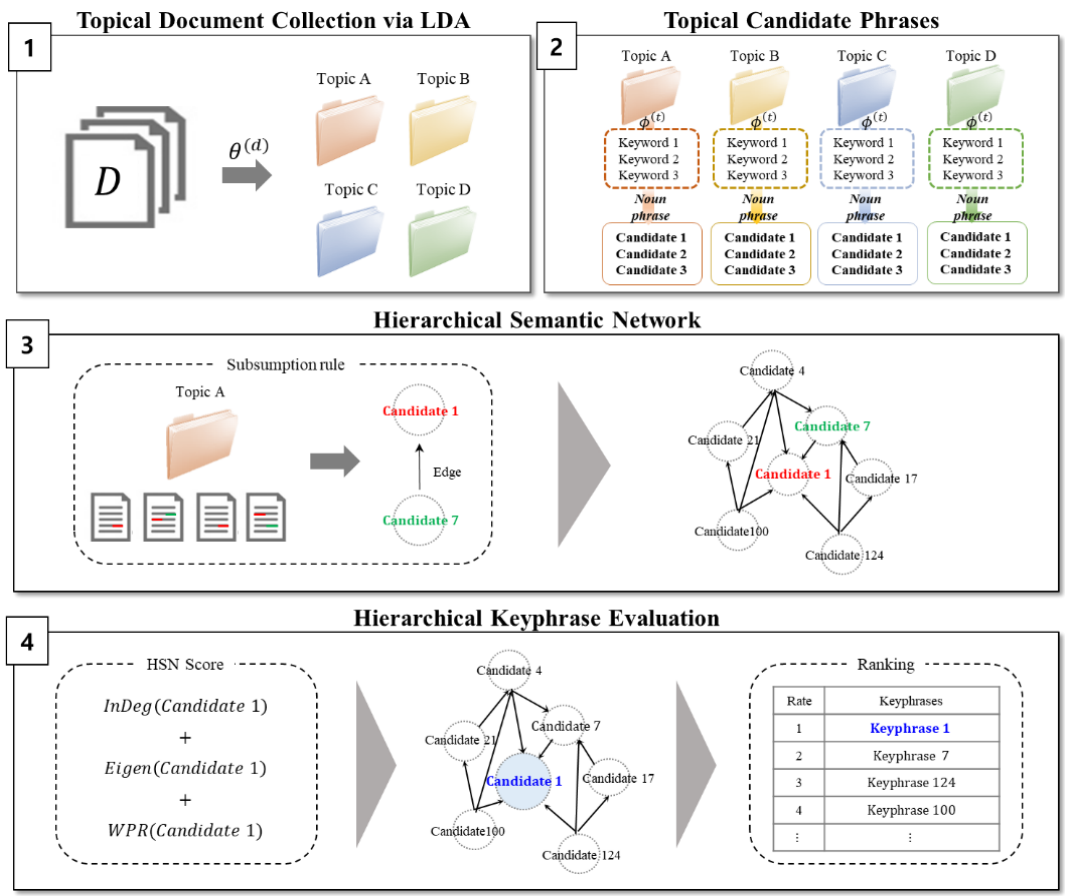

Fig. 1. Overview of the proposed HSN

### 3.2. Topical document collection by LDA

To obtain topical keyphrases, we collect the documents associated with the main topics from document data (i.e., topical document collection) and then extract topical keyphrases from each of them. We use LDA to collect the documents in topics. In LDA, the documents are regarded as random mixtures of latent topics that are characterized by distribution over all the words. Each word $w$ of document $d$ is generated by first sampling a topic $t$ from the corresponding topic distribution per document ($\theta^{(d)}$) and then sampling a word from the corresponding word distribution per topic ($\phi^{(t)}$). To cover a document, this process is repeated for every word in the document. $\theta^{(d)}$ and $\phi^{(t)}$ are parameterized by conjugate Dirichlet priors $\alpha$ and $\beta$. The word probability of $w$ is represented as follows:

$$P(w|d,\alpha,\beta) = \sum_{t=1}^{T} P(w|\phi^{(t)},\beta)P(t|\theta^{(d)},\alpha), \quad (1)$$

where the number of topics is given by $T$ [9, 12]. In the proposed method, we set $\alpha$ as an asymmetric prior from the corpus for $\theta^{(d)}$ because this induces a more robust and data-driven model than a symmetric prior [15]. Thus, we obtain the per-document topic distribution $\theta^{(d)}$ and the per-topic word distribution $\phi^{(t)}$. Both are used in the proposed method. After the per-document topic distribution has been determined, the documents are collected by topic $t$ as follows:

$$C_t = \{d : \theta_t^{(d)} \geq \tau\}, \quad (2)$$

where $\tau$ is a threshold for gathering the documents most related to topic $t$. $C_t$ is the document collection for topic $t$ where $\theta_t^{(d)}$ is larger than $\tau$.

### 3.3. Topical candidate phrase composition

From each topical document collection, we extract the most representative candidate phrases. From the LDA model used to obtain the topical document collections, we use $\phi^{(t)}$ to select the 30 words with the highest probability for each topic. These are selected as LDA keywords for topic $t$. For every $C_t$, we retrieve all the noun phrases that contain the LDA keywords for $t$. Noun phrases are defined as "Adjective+Noun," "Noun+Noun," and "Adjective/Noun+Noun" forms. Through a linguistic processing step, we identify each word in $C_t$ using part-of-speech tagging [16] and obtain the noun phrases. We discard any noun phrase that includes no LDA keywords. We assume the use of 30 words is generally acceptable because using more words would include redundant information in the topic representation. Noun phrases are specifically selected on the premise that they convey most of the meaning of a text [17]. These are selected as candidate phrases for every topic $t$ and are used to construct the HSN in the next section.

### 3.4. Construction of a hierarchical semantic network

Among the candidate phrases for topic $t$, we attempt to select the most comprehensive and representative. It is conceivable that the hierarchical organization of the candidate phrases would facilitate this selection. The principles of hierarchical organization of phrases are as follows: (1) Every phrase in the hierarchy is drawn from the documents and best reflects the topics covered there. (2) The hierarchical organization would be such that an antecedent phrase would refer to a more general concept than its consequent phrase. The concept of antecedent subsumes the concept of consequent [10].

Considering these principles, the practicalities of constructing an HSN for each topic are addressed in the proposed method. The proposed HSN uses candidate phrases that best reflect the topics as components of the network. Moreover, the subsumption relationships of candidate phrases for each topic define the nodes, edge directions, and edge weights of the networks.

When constructing an HSN, we consider a weighted directed network $G := (V, E)$ with node set $V \triangleq \{1, ..., n\}$, edge set $E \subseteq V \times V$, and weighted matrix $H = [h_{s,u}]$. An edge $(s, u) \in E$ denotes that node $u$ obtains information from node $s$, but not vice versa. The adjacency matrix $A = [a_{s,u}] \in \mathrm{R}^{n \times n}$ associated with $G$ is defined as $a_{s,u} > 0$ if $(u, v) \in E$ and $a_{s,u} = 0$ otherwise [18]. It represents edge directions for connected nodes. The weight matrix $H = [h_{s,u}]$ associated with $E$ is obtained from a modified adjacency matrix composed of edge weights [19].

### 3.4.1. Identifying the nodes

Before determining the nodes, we select only those phrases likely to have a hierarchical relationship. We thus dismiss any redundant phrases unrelated to other phrases in the document. We assumed that phrase pairs that co-occur in the same document are semantically related. To incorporate this idea, the proposed HSN adopts the "support" metric of the association rule, which indicates the proportion of the documents in $C_t$ in which a

phrase pair appears [20]. Specifically, the support of phrase pair $X$ can be defined as follows:

$$supp(X:C_t) = \frac{|\{d \in C_t: X_1, X_2 \subseteq d\}|}{|C_t|}, \quad (3)$$

where $X$ is a phrase pair from $Ppairs_t$ (which is the set of all possible pairs of candidate phrases for topic $t$), $X_1$ and $X_2$ denote phrases in the pair $X$, and $|C_t|$ is the number of documents in $C_t$. Then, we select those phrase pairs whose support exceeds the user-specified threshold. That is,

$$S_t = \{X \in Ppairs_t: supp(X:C_t) \geq \eta\}, \quad (4)$$

where $\eta$ is the support threshold. The phrase pairs in $S_t$ correspond to the nodes in $V$ of $G := (V, E)$ for topic $t$. The implication is that a pair with large support, that is, two phrases that frequently appear simultaneously in a document, has a relatively strong semantic connection.

3.4.2. Defining the edges

After the phrase pairs of $S_t$ have been used to define the nodes of the network, the proposed method identifies $E$ (edges and edge directions) of $G := (V, E)$ through the subsumption rule. The nodes may be either antecedent or consequent, and the edges are directed from consequents to antecedents [10]. According to the association rule, the confidence of a phrase pair can be defined as follows [21]:

$$conf(X) = \frac{|\{d \in C_t: X_1 \in d, X_2 \in d\}|}{|\{d \in C_t: X_1 \in d\}|}, \quad (5)$$

where $X$ denotes a phrase pair $\{X_1, X_2\}$. Specifically, confidence is the proportion of documents that contain both $X_1$ and $X_2$ among the documents that contain only $X_1$ in $C_t$. This implies the probability of the presence of $X_2$ when $X_1$ is present. Because the "confidence" metric of the association rule involves the same idea, Equation (5) can be used to assess the subsumption relationships of phrases as follows:

$$conf(X) < 1 \text{ and } conf(X') = 1, \quad (6)$$

where $X$ denotes a phrase pair composed of $\{X_1, X_2\}$ and $X'$ denotes the same phrase pair in reverse order (i.e., $\{X_2, X_1\}$). If the conditions of Equation (6) are met, $X_1$ is the antecedent and $X_2$ is the consequent. The implication is that $X_1$, which entails more general meaning, always tends to exist when $X_2$, which entails less general meaning, exists in a document. Meanwhile, $X_2$ does not always tend to exist when $X_1$ exists in the document. Thus, the edge between the two phrases is directed from $X_2$ to $X_1$, which implies that $X_1$ contains more general concepts than $X_2$ [10]. Consequently, the edge set $E$ of $G := (V, E)$ for topic $t$ is defined through the subsumption relationship of phrases. The nodes that do not satisfy Equation (6) are discarded from $V$ because their edge directions cannot be determined.

3.4.3. Determining the edge weights

To complete the construction of the HSN, we use the metric "lift," which determines edge weights based on the degree of relationship of two connected nodes. Lift can quantify the subsumption relationship between two phrases; greater lift values indicate stronger relationships because lift refers to the dependency probability of two phrases [20]. When two events are independent of each other, it would be meaningless to draw any relationship from them. The lift of the phrase pair $X$ can be defined as follows:

$$lift(X) = \frac{supp(X_1, X_2)}{supp(X_1) supp(X_2)}, \quad (7)$$

and measures the degree to which phrase $X_1$ subsumes phrase $X_2$. It is used to define $H = [h_{s,u}]$.

3.5. Evaluating candidate phrases

Once we have constructed the HSN for each topic, we measure and rate each phrase through indegree centrality, eigencentrality, and WPR in directed networks [19, 22]. We select these three measures among other centrality network measures because they primarily use inflowing edges and directions for properly measuring antecedent nodes in the network. Other measures such as betweenness centrality weigh the salience of a node by its transferability in the shortest path [23]. The followings are definitions and brief explanations of the three

measures. The indegree centrality score of node $u$ in a directed network can be calculated as [24, 25]

$$InDeg(u) = \sum_{s \in V, s \neq u} a_{s,u}. \quad (8)$$

That is, indegree centrality sums the number of inflowing edges [22]. Therefore, the node with the largest number of inflowing edges is granted the highest score. In the HSN, the phrase that contains the most subsuming information is likely to receive the highest score because the consequent phrase is directed to the antecedent phrase.

In addition, we use eigencentrality score of node $u$, which can be calculated as [25, 26]

$$Eigen(u) = \frac{1}{\lambda} \sum_{s \in V} a_{s,u} Eigen(s), \quad (9)$$

where $\lambda$ is the eigenvalue of the adjacency matrix $a_{s,u}$. Eigencentrality measures the salience of nodes that are recursively related to directly connected nodes [27]. For example, in the HSN, node $s$, which is determined to be representative of a topic, can contribute to the representativeness of node $u$. The application of eigencentrality to the HSN is particularly meaningful because it can globally reflect the hierarchical relationships of phrases.

Finally, we use WPR, which, in addition to edge direction, considers edge weights, when scoring a node. We identify the degree of relationship of candidate phrases using the lift values in Equation (8). The WPR score of node $u$ is calculated as [19]

$$WPR(u) = (1-d) + d \sum_{s \in V} WPR(s) H_{s,u}^{in} H_{s,u}^{out}, \quad (10)$$

where $d$ is the damping factor. $H_{s,u}^{in}$ is the weight of $edge(s, u)$, calculated based on the number of inflowing edges of node $u$ and the number of inflowing edges of all nodes $s$. $H_{s,u}^{out}$ is the weight of $edge(s, u)$ calculated based on the number of outflowing edges of node $u$ and the number of outflowing edges of all nodes $s$. By using WPR, we can estimate the degree of generality or representativeness of the antecedent phrase with respect to its consequent nodes.

After computing the node scores in terms of indegree centrality, eigencentrality, and WPR, we scale each network measure score from zero to one to obtain the scaled score of each node with respect to the other nodes. This is the degree of representativeness of a specific phrase compared with the other phrases. We then select the top $R$ phrases according to the three measures and form the set $S_r$ as follows:

$$S_r = \{u \in V : r_{InDeg}(u) \leq R, r_{Eigen}(u) \leq R, r_{WPR}(u) \leq R\}, \quad (11)$$

where $r_{InDeg}(u), r_{Eigen}(u),$ and $r_{WPR}(u)$ refer to the ranks of phrases with respect to each measure. From $S_r$, we select the phrases that are present in all three measure results and assign it as HSN-Keyphrase. If a phrase is not present in all three measure results, we discard the phrase. Consequently, we consider only the relatively highly ranked phrases with respect to all three network measures.

The following HSN-Score is calculated by summing the scaled scores of the corresponding HSN-Keyphrase.

$$HSN - Score(u) = \begin{cases} InDeg(u) + Eigen(u) + WPR(u) & if\ u \in S_r \\ 0 & otherwise \end{cases}, \quad (12)$$

where $InDeg(u), Eigen(u),$ and $WPR(u)$ are the scores of the phrase (node) $u$. We limit the keyphrases as in Equation (12), to prevent the bias of a network measure score from excessively influencing the HSN-Score. For example, without using $R$, the HSN-Score of the HSN-Keyphrase may be excessively high when the indegree score of a node is overly high and the other two scores are overly low. Finally, by ranking the HSN-Keyphrase using the sorted HSN-Keyscore, we obtain the keyphrases by the proposed scheme. We perform the same iterative process for all candidate phrases in each topic.

## 4. Experimental evaluation
### 4.1. Experimental data

To evaluate the performance of the proposed HSN, we used two datasets from the Scopus and Arxiv databases, which include scientific papers, books, and conference proceedings.

Both datasets concern artificial intelligence, computer science, and machine learning. The Scopus dataset consists of 108,200 abstracts, and the Arxiv dataset consists of 13,176 abstracts. We tested the algorithms in different topical document collections obtained from LDA models using different numbers of topics $T$. The topical keyphrase extraction results could vary, depending on $T$, because the topical keyphrases were generated from the phrases in each topical document collection. We set $T$ to 10, 20, and 30. We speculated that if the proposed method outperformed other keyphrase extraction methods regardless of $T$, this would improve its robustness. We set $\tau$, the threshold for generating the topic-document collection, to 0.95, 0.90, and 0.80. We set the minimum support threshold ($\eta$) to 0.005 to obtain the candidate phrase pairs.

### 4.2. Keyphrase extraction results of the proposed method

This section presents the results of the proposed method tested on the Scopus and Arxiv datasets. Because of space limitations, only the results for Topic 1 of $T = 10$ for each dataset are presented. We describe the construction of the HSN for each topic, the ranking and scaling of phrases with respect to different centrality measures, and the final keyphrase extraction. Fig. 2 shows an example of phrase pairs and edges for Topic 1 of $T = 10$ from the Scopus and Arxiv datasets. We note that Fig. 2 partly presents the complicated hierarchical semantic network to exemplify how the phrase pairs are connected to construct the network. The nodes refer to phrases, the edge arrows are directions that indicate which phrases are antecedents or consequents, and the edge weights are degree to which the linked phrases are related. After determining the appropriate candidate phrase pairs, edge directions, and edge weights, we constructed the HSN for each topic.

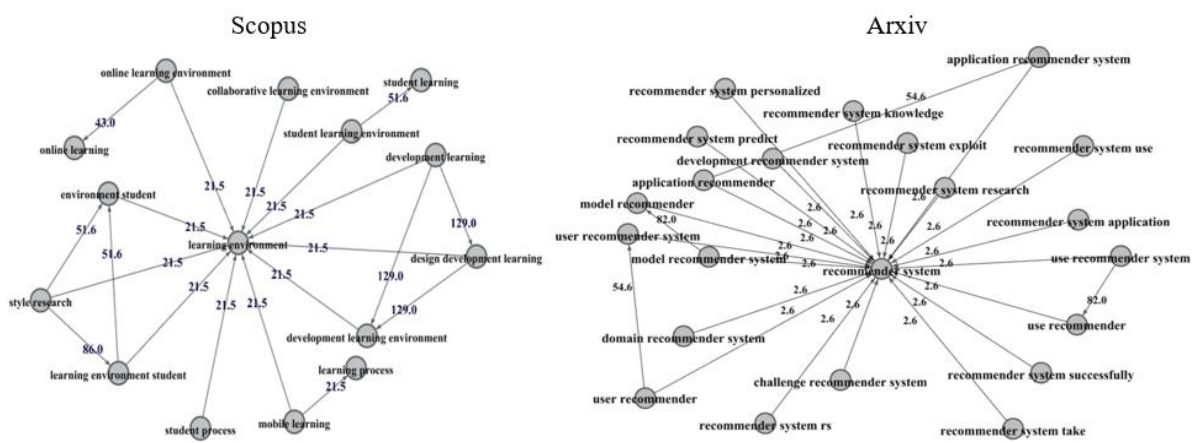

Fig. 2. Example of partial hierarchical semantic networks for Topic 1 of $T = 10$ from the Scopus and Arxiv datasets

Subsequently, we rated each phrase using the three network measures, namely, indegree centrality, eigencentrality, and WPR. After separately computing the node scores in terms of indegree centrality, eigencentrality, and WPR, we scaled each network measure score from zero to one. Then, we selected the top $R$ phrases with respect to the three measures. We set $R$ to 100 to obtain unbiased results when selecting the HSN-Keyphrase from the network measures. Table 1 shows an example of the network measure results.

From the top $R$ phrases, we selected the phrase that simultaneously maximizes the three scores and used it as the HSN-Keyphrase. Moreover, the HSN-Score was calculated by summing the scaled scores of the corresponding HSN-Keyphrase. Table 2 shows an example of the final keyphrase extraction results. For every topic in the document collections obtained using a different number of topics ($T$ = 10, 20, 30), we selected the 10 most highly ranked topical keyphrases by each algorithm.

Table 1. Example of the centrality network measure results for Topic 1 of $T$ = 10 from the Scopus and Arxiv datasets

| | Scopus | | | | | |
|---|---|---|---|---|---|---|
| | **Indegree** | | **Eigencentrality** | | **WPR** | |
| Rank | Candidate Phrase | Score | Candidate Phrase | Score | Candidate Phrase | Score |
| 1 | human perspective | 1.00 | human perspective | 1.00 | human perspective | 1.00 |
| 2 | agent human | 1.00 | agent human | 1.00 | learning environment | 0.80 |
| 3 | generative level dialogue | 0.89 | dialogue model dialogue | 0.83 | social medium | 0.80 |
| 4 | manager dedicated agent | 0.79 | generative level dialogue | 0.69 | dialogue model dialogue | 0.77 |
| | Arxiv | | | | | |
| | **Indegree** | | **Eigencentrality** | | **WPR** | |
| Rank | Candidate Phrase | Score | Candidate Phrase | Score | Candidate Phrase | Score |
| 1 | recommender system | 1.00 | recommender system | 1.00 | recommender system | 1.00 |
| 2 | recommendation system | 0.48 | recommendation system | 0.44 | recommendation system | 0.17 |
| 3 | user item | 0.24 | user item | 0.22 | online learning | 0.08 |
| 4 | large scale | 0.21 | large scale | 0.21 | social network | 0.07 |

Table 2. Example of HSN-Keyphrases and HSN-Scores for Topic 1 of $T$ = 10 from the Scopus and Arxiv datasets

| | Scopus | | Arxiv | |
|---|---|---|---|---|
| Rank | HSN-Keyphrase | HSN-Score | HSN-Keyphrase | HSN-Score |
| 1 | human perspective | 3.00 | recommender system | 3.00 |
| 2 | social networking | 2.09 | recommendation | 1.10 |
|   |                       |      | system              |      |
| 3 | human behavior        | 1.77 | user item           | 0.53 |
| 4 | learning environment  | 1.35 | user preference     | 0.27 |
| 5 | social medium         | 1.18 | online learning     | 0.16 |
| 6 | collaborative learning| 0.79 | base recommendation | 0.16 |
| 7 | social science        | 0.64 | item user           | 0.16 |
| 8 | social network        | 0.56 | recommendation user | 0.14 |
| 9 | college student       | 0.47 | information user    | 0.14 |
| 10| university student    | 0.35 | user profile        | 0.13 |

4.3. Experimental Evaluation

We performed a blind test to assess the validity of the proposed method [25]. We used Amazon Mechanical Turk (Mturk), an online web-based platform (https://www.mturk.com), for conducting surveys and recruiting randomly selected people to take the blind test. Amazon Mturk is a reliable survey source because its participants are believed to be sufficiently representative and diverse to be involved in research studies [28, 29]. The participants in the blind test received the titles most related to the topic as a guideline. One hundred people chosen at random participated in ranking the six algorithms according to their satisfaction with the topical keyphrase sets. Their satisfaction depended on how well each keyphrase set represented and summarized the topic. Among the six algorithms, single-keyword extraction methods (LDA, TF-ITF) were included to globally evaluate the proposed method. For every topic in the document collections obtained using a different number of topics ($T = 10, 20, 30$), we averaged the rankings of the six algorithms, as determined by the participants. The lowest value represents the highest rank.

The blind test was designed using documents highly related to the topic and six topical keyphrase sets. We selected five documents that had the highest probability of containing the topic from the LDA model and gathered their titles. For instance, an example title set for Topic 1 of $T = 10$ from the Scopus dataset is as follows: (1) "Differences between adaptors and innovators in the context of entrepreneurial potential dimensions," (2) "Research for finding relationships between mass media and social media based on agenda setting theory," (3) "Identification of the learning behavior of students for education personalization," (4) "With a little help from my friends: a computational model for the role of social support in mood regulation," and (5) "Lecturer perceptions of impoliteness and inappropriateness in student e-mail requests: a Norwegian perspective." These titles were provided to the participants, who were asked to rate the six topical keyphrase extraction methods. Table 3 shows an example of the six topical keyphrase extraction results.

Table 3. Example of topical keyphrase extraction results for Topic 1 of $T = 10$ from the Scopus dataset

| Rank | LDA    | TF–ITF  | N-gram TF–ITF | TR      | TPR            | HSN (proposed)    |
|------|--------|---------|---------------|---------|----------------|-------------------|
| 1    | social | student | social medium | student | social network | human perspective |
| 2 | model | social | social network | modelling | social influence | social networking |
| 3 | system | teacher | learning environment | modeling | social community | human behavior |
| 4 | paper | education | college student | modeler | social study | learning environment |
| 5 | datum | school | digital game | social | social medium | social medium |
| 6 | user | educational | mass medium | paper | social dynamic | collaborative learning |
| 7 | behavior | game | high school | researcher | social behavior | social science |
| 8 | human | child | entrepreneurial potential | researched | social analysis | social network |
| 9 | study | cultural | social interaction | study | social interaction | university student |
| 10 | student | agent | collaborative learning | educational | social activity | college student |

An example title set for Topic 1 of $T = 10$ from the Arxiv dataset is as follows: (1) "Towards effective research-paper recommender systems and user modeling based on mind maps," (2) "Communications and control for wireless drone-based antenna array," (3) "Intent-aware contextual recommendation system," (4) "Probabilistic graphical models for credibility analysis in evolving online communities," and (5) "Explainable recommendation: theory and applications." These titles were provided to the participants to rate the six topical keyphrase extraction methods. Table 4 shows an example of the six topical keyphrase extraction results.

Table 4. Example of topical keyphrase extraction results for Topic 1 of $T$ = 10 from the Arxiv dataset

| Rank | LDA | TF–ITF | N-gram TF–ITF | TR | TPR | HSN (proposed) |
|---|---|---|---|---|---|---|
| 1 | user | recommendation | recommender system | user | neural network | recommender system |
| 2 | time | session | user preference | recommend | deep network | recommendation system |
| 3 | social | item | social dilemma | recommender | convolutional network | user item |
| 4 | model | user | recommendation system | recommendable | present network | user preference |
| 5 | system | email | user item | model | propose method | online learning |
| 6 | use | recommender | collaborative filtering | modeling | present method | base recommendation |
| 7 | network | venue | cold start | use | specific network | item user |
| 8 | analysis | rating | user review | useful | deep neural | recommendation user |

| 9 | base | cilantro | point cloud | preference recommendation | different feature | information user |
| 10 | datum | ranker | expert user | base | small network | user profile |

After obtaining the final keyphrase extraction results of the proposed HSN, we compared the six topical keyphrase results for a different number of topics ($T$ = 10, 20, 30). This demonstrated that the extracted keyphrases by the proposed HSN were better than those by the other five algorithms in various settings. We averaged the ranks by the participants for each topic. (See Appendix A for details.) To facilitate the interpretation of the results, we averaged the average ranks for each topic. Fig. 3 shows the average of the average ranks of the six algorithms in terms of their accuracy with different numbers of topics in the Scopus dataset.

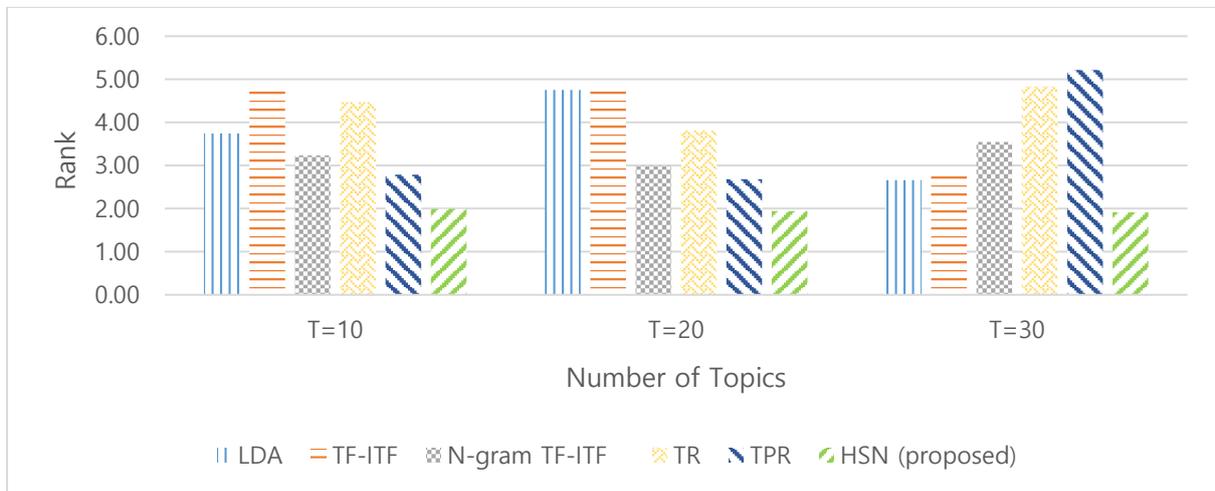

Fig. 3. Performance comparison of LDA, TF–ITF, N-gram TF–ITF, TR, TPR, and HSN for different numbers of topics ($T$ = 10, 20, 30) from Scopus dataset

The overall user satisfaction results demonstrate that the proposed HSN selected the most meaningful keyphrases to represent the topics. Because the LDA and TF–ITF are primarily unigram-centric, their results have limited semantic capacity. For example, "social" in the TF–ITF results in Table 3 is insufficient to represent Topic 1 of the Scopus dataset. Moreover, although the N-gram-based TF–ITF algorithm ensures the "semantical richness" of the extracted keyphrases, it suffers from some inherent problems that result in its inability to detect the most representative topical keyphrases. Both TF–ITF algorithms depend mostly on phrase frequency. For example, in Table 3, "college student" is ranked fourth in the N-gram-based TF–ITF results, but "collaborative learning" is ranked tenth. Obviously, the keyphrase "collaborative learning" is a more general concept than "college student" is, and therefore it should be regarded as the main theme of the entire context. It is conceivable that a superior topical keyphrase extraction method should distinguish the two keyphrases and select "collaborative learning." Therefore, the results of TF–ITF algorithms are often biased, and higher phrase frequency favors them.

The performance of TR is also worse than that of the proposed HSN. Because the algorithm assumes that relevant phrases are neighbors, it creates adjacent network edges.

This sequence information derives a strong relevance only between nearby words and hence reflects only a local portion of the document. When PageRank is applied to the network, it outputs only similar nearby words. Consequently, the results are narrow and unrepresentative of the full scope of the documents. For example, in Table 3, TR extracts "modeling," "modelling," and "modeler" as its top-ranked results. However, these three words appear to have been extracted only because they are lexically distinct, and no extraction was made for keyphrases that globally represent the entire topic. Moreover, TR cannot distinguish hierarchical relationships among the phrases; therefore, it yields keyphrases that are related to a redundant concept such as "model."

The performance of TPR is also considered poor compared with that of the proposed HSN. As in the case of TR, the edge weights of the nodes (phrases) are based on phrase co-occurrence within a word sequence window. This prevents the algorithm from extracting global keyphrases related to the topic. Moreover, TPR adds the word probability for a topic from the LDA to the PageRank scoring formula. Directly adding such a term to the scoring function may reduce the accuracy of the PageRank scoring formula because the cumulative sum of phrase scores is calculated. Thus, when the per-topic word probability is explicitly added to the scoring formula, the score of specific keyphrases is exponentially increased, yielding biased results. For example, in Table 3, the most highly ranked word of the LDA results is "social." Because the TPR is directly influenced by the probability of the word "social," its results all include it. Such results are unrepresentative of the entire topic. By contrast, the results do not include keyphrases such as "human perspective" and "learning environment." This implies that excessive weighting of a word probability term can hamper a ranking algorithm in its selection of the most general concepts.

Fig. 4. shows the average ranks of the six algorithms with different number of topics ($T$=10, 20, 30) in terms of their accuracy in the Arxiv dataset. We averaged the ranks by the participants. (See Appendix A for details.) To facilitate the interpretation of the results, we averaged the average ranks for each topic. Fig.4. shows the average of the average ranks of the six algorithms in terms of their accuracy for different numbers of topics in the Arxiv dataset.

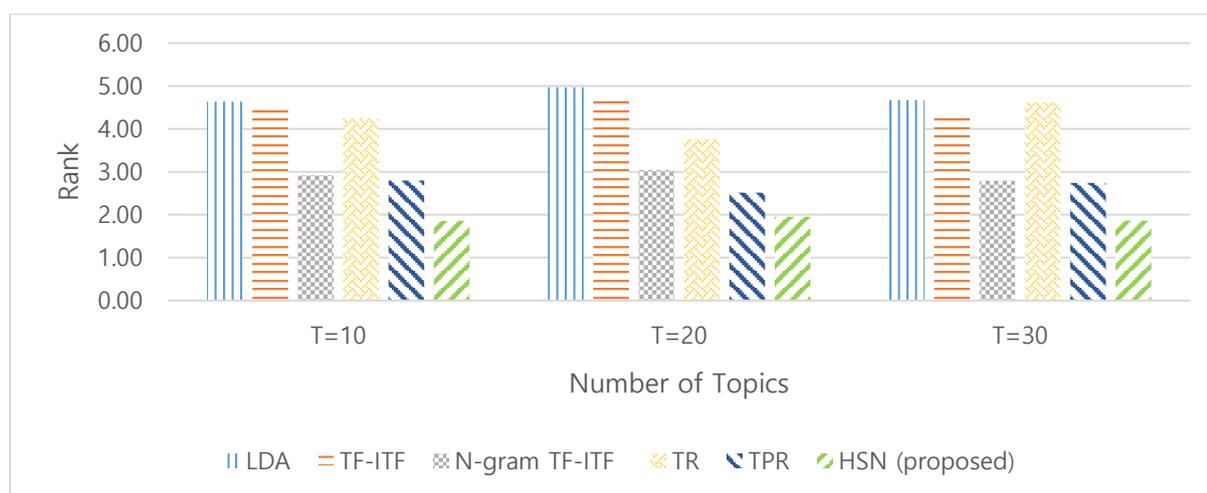

Fig. 4. Performance comparison of LDA, TF–ITF, N-gram TF–ITF, TR, TPR, and HSN for different numbers of topics ($T = 10, 20, 30$) from the Arxiv dataset

The overall user satisfaction results demonstrate that the proposed HSN selected the most appropriate keyphrases to represent the topics. The analysis of the Arxiv dataset results for different numbers of topics are in line with those of the Scopus dataset. For example, in the LDA and TF-ITF results in Table 4, keywords such as "user" and "recommendation" do not carry sufficient information to represent Topic 1 of the Arxiv dataset. Moreover, N-gram TF–ITF keyphrase results, such as "social dilemma" and "cold start," have no content that can summarize Topic 1. This suggests that considering only phrase frequency may prevent the extraction of the most representative phrase from the topical document collection. Although TR and TPR yield results that reflect the semantic relationships between phrases, they adopt non-hierarchical relationships. For example, in Table 4, the TR produces several redundant words, such as "model," "modeling," "use," "useful," "base," and "user," as its top extracted results. These words not only have limited semantic capacity but also lack the lexical abundance to be regarded as representations of the topical documents.

It is conceivable that a keyphrase extraction model should properly select the core words that can best represent the topic, and it should discard the others. However, the results of TPR in Table 4 do not contain any critical information that can summarize the topic; rather, there are numerous phrases including the word "network." This bias seems to derive from the modified PageRank scoring formula, which is explicitly influenced by the probability of the word "network." Thus, as in the Scopus dataset, the proposed HSN exhibits superior performance compared with the other algorithms in the Arxiv dataset.

## 5. Conclusions

Manually organizing a vast number of collected documents and determining their main theme is a tedious task. Topical keyphrases can serve as guidelines so that users may briefly overview the content of such a collection. Several studies have been conducted to develop an efficient topical keyphrase extraction method; however, existing methods use simple term frequency rules and nonsemantic findings. To overcome this limitation, we use the HSN, which allows us to quantify direct, indirect, and integral relations among topical keyphrases [25, 30].

From the experimental results and analysis, we highlighted the following findings and practical implications. The underlying principles of TF–ITF and N-gram-based TF–ITF consider only term frequency from a topical bag-of-words without forming any semantic relationships. Thus, they are unlikely to capture the general meaning of a text. The proposed method overcomes this limitation by extracting N-gram keyphrases and using a hierarchical semantic network. Consequently, the most representative phrases can be identified. The major limitation of TR and TPR is the use of a term counting processes in a word sequence of limited size. Such proximity relationships produce meaningless phrases. On the contrary, the proposed HSN emphasizes the hierarchical semantics of each keyphrase using three network measures. Thus, the proposed HSN ensures the quality, diversity, and representativeness of the extraction results, which can well represent the given documents.

In future work, we would like to apply the topical keyphrase extraction methods to document summarization tasks. Because the topical keyphrase results capture the most general concept of each topic, it would be interesting to aggregate the keyphrases into valid sentences. The sentences could then be compounded to become a summary. Moreover, we would like to assign more weight to a specific network measure when detecting semantic relationships. As these approaches confer different advantages, meticulously applying them could certainly improve the proposed method.

## Acknowledgements

The authors would like to thank the editor and reviewers for their useful comments and suggestions, which were greatly help in improving the quality of the paper. This research was supported by Brain Korea PLUS, the Ministry of Trade, Industry & Energy under Industrial Technology Innovation Program (R1623371) and the Institute for Information & Communications Technology Promotion grant funded by the Korea government (No. 2018-0-00440, ICT-based Crime Risk Prediction and Response Platform

Development for Early Awareness of Risk Situation), and the Ministry of Culture, Sports and Tourism and Korea Creative Content Agency in the Culture Technology Research & Development Program 2019.**References**

[1] V. S. Anoop, S. Asharaf, P. Deepak, Unsupervised concept hierarchy learning: a topic modeling guided approach, Procedia Comput. Sci. 89 (2016) 386–394.

[2] M. Zihayat, A. Ayanso, X. Zhao, H. Davoudi, A. An, A utility-based news recommendation system, Decis.Support Syst. 117 (2019) 14–27.

[3] M. Scholz, C. Brenner, O. Hinz, AKEGIS: automatic keyword generation for sponsored search advertising in online retailing, Decis.Support Syst. (2019).

[4] T. Tomokiyo, M. Hurst, A language model approach to keyphrase extraction, Proceedings of the ACL 2003 workshop on Multiword expressions: analysis, acquisition and treatment. 2003.

[5] K.S. Hasan, V. Ng, Automatic Keyphrase Extraction: A Survey of the State of the Art, Proceedings of the 52nd Annual Meeting of the Association for Computational Linguistics, 1, 2014, pp. 1262–1273.

[6] Z. Liu, P. Li, Y. Zheng, M. Sun, Clustering to find exemplar terms for keyphrase extraction, Proceedings of the 2009 Conference on Empirical Methods in Natural Language Processing, Association for Computational Linguistics, 2009, pp. 257–266.

[7] Y. Shynkevich, T.M. McGinnity, S.A. Coleman, A. Belatreche, Forecasting movements of health-care stock prices based on different categories of news articles using multiple kernel learning, Decis. Support. Syst. 85 (2016) 74–83.

[8] G. Li, H. Wang, Improved automatic keyword extraction based on TextRank using domain knowledge, In Natural Language Processing and Chinese Computing, Springer, Berlin, Heidelberg, 2014, pp. 403–413.

[9] Z. Liu, W. Huang, Y. Zheng, M. Sun, Automatic keyphrase extraction via topic decomposition, In Proceedings of the 2010 conference on empirical methods in natural language processing, 2010 (October), pp. 366–376.

[10] M. Sanderson, B. Croft, Deriving concept hierarchies from text, MASSACHUSETTS UNIV AMHERST DEPT OF COMPUTER SCIENCE, 2005.

[11] C. Wang, S. Li, CoRankBayes: Bayesian learning to rank under the co-training framework and its application in keyphrase extraction, In Proceedings of the 20th ACM international conference on Information and knowledge management, ACM, 2011 (October), pp. 2241–2244.

[12] D.M. Blei, A.Y. Ng, M.I. Jordan, Latent dirichlet allocation, J. Mach. Learn. Res. 3 (2003) 993–1022.

[13] R. Mihalcea, P. Tarau, Textrank: Bringing order into text, In Proceedings of the 2004 conference on empirical methods in natural language processing, 2004.

[14] A. Bellaachia, M. Al-Dhelaan, Ne-rank: A novel network-based keyphrase extraction in twitter, In Proceedings of the The 2012 IEEE/WIC/ACM International Joint Conferences on Web Intelligence and Intelligent Agent Technology, IEEE Computer Society, 2012, pp. 372–379.

[15] H.M. Wallach, D.M. Mimno, A. McCallum, Rethinking LDA: Why priors matter, In Advances in neural information processing systems, 2009, pp. 1973–1981.

[16] E. Brill, A simple rule-based part of speech tagger, In Proceedings of the third conference on Applied natural language processing, Association for Computational Linguistics, 1992 (March), pp. 152–155.

[17] A.A. Freihat, B. Dutta, F. Giunchiglia, Compound noun polysemy and sense enumeration in wordnet, In Proceedings of the 7th International Conference on Information, Process, and Knowledge Management (eKNOW), 2015, pp. 166–171.

# Appendix

Tables A1, A2, and A3 show the average ranking that the participants assigned to the six algorithms in terms of their accuracy in representing the topics in the Scopus dataset. Boldface entries represent the highest average rank awarded by the 100 participants, and the values in the brackets indicate the standard deviation of the ranks. For statistical comparison, we conducted the Wilcoxon Rank-sum test (a nonparametric test) to compare the ranks. We rejected the null hypothesis that the ranks of each algorithm and the HSN do not significantly deviate if the p-value was lower than 0.05.

Table A1. Comparison of six topical keyphrase extraction algorithms. Each value indicates the average rank by the participants for $T = 10$ in the Scopus dataset. The average rank of each algorithm is listed in the bottom row. Each value in a bracket indicates the standard deviation of the ranks. Each * tagged on the average rank represents statistically significant difference of the ranks between the corresponding algorithm and the proposed HSN, with the p-value being lower than 0.05

| Topic | LDA | TF–ITF | N-gram TF–ITF | TR | TPR | **HSN (proposed)** |
|---|---|---|---|---|---|---|
| 1 | 3.36* (1.44) | 4.42* (1.47) | 3.65* (1.36) | 4.99* (1.26) | 2.30* (1.33) | **2.27** (1.40) |
| 2 | 3.94* (1.30) | 4.96* (1.19) | 3.06* (1.44) | 4.62 (1.23) | 2.54* (1.45) | **1.88** (1.17) |
| 3 | 2.72* (1.63) | 5.03* (1.38) | 3.59 (1.25) | 4.53* (1.25) | 3.03* (1.29) | **2.10** (1.39) |
| 4 | 3.69* (1.31) | 4.78* (1.22) | 3.08 (1.38) | 4.97* (1.17) | 2.62* (1.45) | **1.86** (1.18) |
| 5 | 3.56* (1.37) | 5.27* (1.02) | 3.02* (1.36) | 4.51* (1.29) | 2.63* (1.44) | **2.00** (1.28) |
| 6 | 3.74* (1.22) | 4.85* (1.20) | 4.72* (1.48) | 3.29* (1.37) | 2.37* (1.32) | **2.03** (1.37) |
| 7 | 4.77* (1.17) | 4.97* (1.35) | 2.38* (1.35) | 3.88* (1.14) | 3.03* (1.35) | **1.97** (1.29) |
| 8 | 5.09* (1.18) | 4.60* (1.39) | 3.08* (1.27) | 3.93* (1.21) | 2.46* (1.35) | **1.84** (1.15) |
| 9 | 3.31* (1.36) | 4.94* (1.32) | 3.51* (1.37) | 4.92* (1.08) | 2.26* (1.21) | **2.06** (1.30) |
| 10 | 3.26* (1.40) | 3.88* (1.20) | 2.27* (1.31) | 5.07* (1.14) | 4.62* (1.28) | **1.90** (1.20) |
| Average Rank | 3.74 | 3.88 | 3.24 | 4.47 | 2.79 | **1.99** |

Table A2. Comparison of six topical keyphrase extraction algorithms. Each value indicates the average rank by the participants for $T = 20$ in the Scopus dataset. The average rank of each algorithm is listed in the bottom row. Each value in a bracket indicates the standard deviation of the ranks. Each * tagged on the average rank represents statistically significant difference of the ranks between the corresponding algorithm and the proposed HSN, with the p-value being lower than 0.05.

| Topic | LDA | TF–ITF | N-gram TF–ITF | TR | TPR | **HSN (proposed)** |
|---|---|---|---|---|---|---|
| 1 | 5.08* (1.28) | 4.77* (0.97) | 3.26* (1.29) | 3.58* (1.21) | 2.57* (1.55) | **1.74** (1.12) |
| 2 | 5.28* (1.04) | 4.58* (1.17) | 3.02* (1.32) | 3.60 (1.19) | **2.15*** (1.44) | 2.36 (1.45) |
| 3 | 5.03* (1.31) | 3.60* (1.15) | 2.33 (1.40) | 4.89* (1.20) | 3.08* (1.29) | **2.07** (1.23) |
| 4 | 4.55* (1.07) | 5.23* (1.01) | 2.45 (1.60) | 3.81* (1.33) | 2.83* (1.29) | **2.13** (1.31) |
| 5 | 3.64* (1.15) | 4.79* (1.18) | 3.00* (1.26) | 5.24* (1.12) | 2.51* (1.45) | **1.82** (1.02) |
| 6 | 5.35* (0.94) | 4.73* (1.00) | 2.61* (1.51) | 3.70* (1.17) | 2.74* (1.26) | **1.86** (1.16) |
| 7 | 2.61* (1.57) | 5.15* (1.12) | 3.52* (1.22) | 4.79* (1.18) | 2.98* (1.23) | **1.95** (1.22) |
| 8 | 4.79* (1.21) | 5.21* (1.02) | 3.00* (1.18) | 3.68* (1.32) | 2.39* (1.43) | **1.93** (1.11) |
| 9 | 5.19* (1.10) | 4.84* (1.18) | 3.02* (1.16) | 2.42* (1.49) | 3.58* (1.09) | **1.96** (1.32) |
| 10 | 5.36* (0.95) | 4.74* (0.86) | 2.79* (1.23) | 2.79* (1.72) | 3.53* (1.12) | **1.78** (1.08) |
| 11 | 5.25* (1.05) | 4.68* (0.93) | 3.08* (1.27) | 3.80 (1.30) | 2.28* (1.49) | **1.91** (1.13) |
| 12 | 4.70* (1.10) | 5.21* (1.14) | 3.60* (1.13) | 3.10* (1.35) | 2.41* (1.50) | **1.98** (1.28) |
| 13 | 5.25* (1.03) | 4.56* (1.21) | 3.04* (1.31) | 3.82* (1.17) | 2.47* (1.58) | **1.85** (1.06) |
| 14 | 5.36* (1.00) | 4.87* (0.92) | 2.90* (1.22) | 3.52* (1.07) | 2.45* (1.51) | **1.91** (1.20) |
| 15 | 5.27* (1.02) | 4.65* (1.28) | 2.34 (1.46) | 3.77* (1.21) | 2.95* (1.15) | **2.02** (1.24) |
| 16 | 5.20* (1.12) | 4.67* (1.23) | 3.02* (1.24) | 3.74* (1.14) | 2.48* (1.50) | **1.89** (1.21) |
| 17 | 3.81* (1.13) | 4.62* (1.23) | 2.41* (1.42) | 5.22* (1.24) | 3.01* (1.18) | **1.93** (1.25) |
| 18 | 4.91* (1.02) | 4.97* (1.34) | 3.74* (1.19) | 3.06* (1.34) | 2.40* (1.37) | **1.93** (1.22) |
| 19 | 3.66* (1.12) | 5.16* (1.21) | 3.09* (1.29) | 4.74* (1.26) | 2.44* (1.45) | **1.91** (1.14) |
| 20 | 4.79* (1.14) | 5.17* (1.10) | 3.74* (1.22) | 2.95* (1.23) | 2.44* (1.53) | **1.92** (1.11) |
| Average Rank | 4.75 | 4.81 | 3.00 | 3.81 | 2.68 | **1.94** |

Table A3. Comparison of six topical keyphrase extraction algorithms. Each value indicates the average rank by the participants for $T = 30$ in the Scopus dataset. The average rank of each algorithm is listed in the bottom row. Each value in a bracket indicates the standard deviation of the ranks. Each * tagged on the average rank represents statistically significant difference of the ranks between the corresponding algorithm and the proposed HSN, with the p-value being lower than 0.05.

| Topic | LDA | TF–ITF | N-gram TF–ITF | TR | TPR | **HSN (proposed)** |
|---|---|---|---|---|---|---|
| 1 | 2.28* (1.29) | 3.16* (1.25) | 3.68* (1.09) | 5.01* (1.01) | 5.05* (1.26) | **1.82** (1.15) |
| 2 | **1.72*** (1.02) | 3.16* (1.23) | 3.75* (1.27) | 4.57* (1.22) | 5.32* (0.97) | 2.49 (1.43) |
| 3 | 2.91* (1.19) | 2.44* (1.43) | 3.66* (1.15) | 4.69* (1.02) | 5.51* (0.74) | **1.79** (1.11) |
| 4 | 2.96* (1.32) | 2.30 (1.44) | 3.55* (1.08) | 4.98* (1.03) | 5.04* (1.21) | **2.15** (1.36) |
| 5 | 2.33* (1.34) | 3.25* (1.24) | 3.85* (1.15) | 4.60* (1.29) | 5.24* (0.98) | **1.74** (1.14) |
| 6 | 2.32* (1.36) | 3.14* (1.36) | 3.54* (1.07) | 4.91* (1.03) | 5.17* (1.06) | **1.91** (1.28) |
| 7 | 3.18* (1.26) | 2.38* (1.48) | 3.61* (1.02) | 4.86* (1.09) | 5.17* (1.05) | **1.81** (1.24) |
| 8 | 2.42* (1.47) | 2.94* (1.27) | 3.67* (1.16) | 4.88* (1.04) | 5.14* (1.14) | **1.95** (1.22) |
| 9 | 3.77* (1.15) | 2.98* (1.17) | 2.34 (1.49) | 4.94* (1.06) | 4.89* (1.42) | **2.07** (1.31) |
| 10 | 2.32 (1.47) | 2.83* (1.28) | 3.68* (1.07) | 4.78* (1.11) | 5.20* (1.05) | **2.19** (1.44) |
| 11 | 2.41* (1.39) | 2.98* (1.30) | 3.61* (1.08) | 4.91* (1.04) | 5.21* (1.12) | **1.88** (1.19) |
| 12 | 2.42* (1.37) | 3.03* (1.13) | 3.69* (1.21) | 4.81* (1.17) | 5.25* (0.99) | **1.79** (1.18) |
| 13 | 2.42* (1.48) | 3.15* (1.30) | 3.43* (0.99) | 4.92* (1.02) | 5.23* (0.90) | **1.86** (1.35) |
| 14 | 2.20* (1.30) | 3.06* (1.28) | 3.80* (1.10) | 4.81* (1.14) | 5.26* (0.89) | **1.88** (1.22) |
| 15 | 2.82* (1.25) | 2.35 (1.42) | 3.84* (1.15) | 4.83* (1.16) | 5.10* (1.19) | **2.06** (1.23) |
| 16 | 2.43* (1.50) | 2.95* (1.14) | 3.75* (1.13) | 4.85* (1.13) | 5.21* (1.00) | **1.80** (1.11) |
| 17 | 3.02* (1.23) | 2.58* (1.52) | 3.70* (1.13) | 4.89* (1.03) | 5.18* (0.96) | **1.63** (1.04) |
| 18 | 3.64* (1.15) | 2.94* (1.27) | 2.35* (1.45) | 4.89* (0.90) | 5.16* (1.06) | **2.03** (1.43) |
| 19 | 2.44* (1.38) | 3.11* (1.29) | 3.59* (1.09) | 4.74* (1.26) | 5.26* (1.09) | **1.86** (1.18) |
| 20 | 3.32* (1.31) | 2.28* (1.37) | 3.46* (1.03) | 4.89* (1.08) | 5.30* (0.89) | **1.75** (1.06) |
| 21 | 2.43* (1.52) | 2.93* (1.24) | 3.65* (1.21) | 4.81* (0.95) | 5.22* (1.12) | **1.94** (1.20) |
| 22 | 3.03* (1.30) | 2.36* (1.22) | 3.72* (1.12) | 4.77* (1.09) | 5.21* (1.24) | **1.93** (1.33) |
| 23 | 2.62* (1.61) | 2.91* (1.19) | 3.46* (1.05) | 4.93* (1.09) | 5.14* (1.14) | **1.93** (1.25) |
| 24 | 2.59* (1.50) | 2.82* (1.13) | 3.73* (1.11) | 4.82* (0.99) | 5.29* (1.03) | **1.75** (1.14) |
| 25 | 2.42* (1.41) | 3.02* (1.29) | 3.60* (1.19) | 4.85* (0.92) | 5.30* (1.05) | **1.81** (1.07) |
| 26 | 3.04* (1.19) | 2.54* (1.52) | 3.48* (1.04) | 4.77* (1.07) | 5.30* (1.18) | **1.88** (1.25) |
| 27 | 2.41* (1.32) | 2.81* (1.20) | 3.56* (1.13) | 4.97* (1.03) | 5.28* (0.96) | **1.97** (1.31) |
| 28 | 2.47* (1.49) | 2.95* (1.24) | 3.55* (1.16) | 4.77* (1.09) | 5.30* (1.03) | **1.96** (1.25) |
| 29 | 2.99* (1.35) | 2.37* (1.36) | 3.56* (1.06) | 4.81* (1.05) | 5.28* (1.02) | **1.98** (1.35) |
| 30 | 2.53* (1.48) | 3.06* (1.26) | 3.50* (1.15) | 4.75* (1.13) | 5.26* (0.99) | **1.90** (1.34) |
| Average Rank | 2.66 | 2.83 | 3.55 | 4.83 | 5.22 | **1.92** |

Tables A4, A5, and A6 show the average rankings that the participants assigned to the six algorithms in terms of their accuracy in representing the topics in the Arxiv data set. Boldface entries represent the highest average rank.

Table A4. Comparison of six topical keyphrase extraction algorithms. Each value indicates the average rank by the participants for $T = 10$ in the Arxiv dataset. The average rank of each algorithm is listed in the bottom row. Each value in a bracket indicates the standard deviation of the ranks. Each * tagged on the average rank represents statistically significant difference of the ranks between the corresponding algorithm and the proposed HSN, with the p-value being lower than 0.05.

| Topic | LDA | TF–ITF | N-gram TF–ITF | TR | TPR | **HSN (proposed)** |
|---|---|---|---|---|---|---|
| 1 | 5.02* (1.33) | 4.07* (1.39) | 2.56* (1.48) | 4.45* (1.25) | 2.89* (1.24) | **2.01** (1.22) |
| 2 | 4.60* (1.38) | 4.98* (1.02) | 3.55* (1.43) | 3.65* (1.26) | 2.50* (1.42) | **1.71** (1.13) |
| 3 | 4.63* (1.49) | 5.03* (1.12) | 3.22* (1.44) | 3.63* (1.26) | 2.65* (1.29) | **1.85** (1.25) |
| 4 | 4.57* (1.22) | 3.83* (1.32) | 2.66* (1.55) | 4.88* (1.39) | 3.13* (1.37) | **1.93** (1.27) |
| 5 | 4.42* (1.30) | 4.15* (1.42) | 2.79* (1.51) | 4.88* (1.20) | 3.14* (1.24) | **1.63** (1.10) |
| 6 | 4.71* (1.31) | 3.89* (1.11) | 2.33 (1.40) | 4.99* (1.18) | 3.07* (1.36) | **2.00** (1.31) |
| 7 | 4.98* (1.11) | 4.25* (1.43) | 2.32* (1.38) | 4.47* (1.23) | 3.00* (1.23) | **1.98** (1.28) |
| 8 | 3.78* (1.14) | 5.11* (1.19) | 3.08* (1.28) | 4.69* (1.19) | 2.59* (1.43) | **1.74** (1.22) |
| 9 | 4.88* (1.28) | 5.01* (1.11) | 3.67* (1.27) | 3.01* (1.15) | 2.57* (1.54) | **1.85** (1.13) |
| 10 | 4.77* (1.32) | 5.11* (0.92) | 3.02* (1.37) | 3.77* (1.21) | 2.45* (1.31) | **1.88** (1.27) |
| Average Rank | 4.64 | 4.54 | 2.92 | 4.24 | 2.80 | **1.86** |

Table A5. Comparison of six topical keyphrase extraction algorithms. Each value indicates the average rank by the participants for $T = 20$ in the Arxiv data set. The average rank of each algorithm is listed in the bottom row. Each value in a bracket indicates the standard deviation of the ranks. Each * tagged on the average rank represent the statistically significant difference of the ranks between the corresponding algorithm and the proposed HSN, with the p-value being lower than 0.05.

| Topic | LDA | TF–ITF | N-gram TF–ITF | TR | TPR | **HSN (proposed)** |
|---|---|---|---|---|---|---|
| 1 | 5.04* (1.31) | 4.81* (1.17) | 3.03* (1.30) | 3.74 (1.18) | 2.31* (1.43) | **2.06** (1.19) |
| 2 | 5.20* (1.10) | 4.70* (1.09) | 3.02* (1.28) | 3.67* (1.38) | 2.61* (1.41) | **1.80** (1.13) |
| 3 | 4.95* (1.30) | 4.75* (1.21) | 3.11* (1.35) | 3.75* (1.30) | 2.43* (1.37) | **2.00** (1.29) |
| 4 | 5.07* (1.12) | 4.64* (1.16) | 3.02* (1.49) | 3.82* (1.43) | 2.52* (1.40) | **1.93** (1.07) |
| 5 | 5.02* (1.24) | 4.74* (1.13) | 3.21* (1.43) | 3.70* (1.34) | 2.48* (1.28) | **1.86** (1.22) |

| Topic | LDA | TF–ITF | N-gram TF–ITF | TR | TPR | HSN (proposed) |
|---|---|---|---|---|---|---|
| 6 | 4.97* (1.19) | 4.67* (1.23) | 3.02* (1.37) | 3.89* (1.37) | 2.56* (1.45) | **1.89** (1.16) |
| 7 | 4.90* (1.29) | 4.80* (1.21) | 3.04* (1.45) | 3.77 (1.38) | 2.33* (1.22) | **2.15** (1.32) |
| 8 | 4.96* (1.17) | 5.03* (1.04) | 3.11* (1.21) | 3.51* (1.25) | 2.56* (1.49) | **1.83** (1.22) |
| 9 | 4.93* (1.31) | 4.78* (1.23) | 2.95* (1.33) | 3.86* (1.36) | 2.58* (1.27) | **1.90** (1.23) |
| 10 | 4.97* (1.25) | 4.81* (1.10) | 3.13* (1.40) | 3.56* (1.27) | 2.61* (1.48) | **1.93** (1.28) |
| 11 | 4.92* (1.33) | 4.88* (1.08) | 2.86* (1.21) | 3.74* (1.36) | 2.54* (1.42) | **2.07** (1.39) |
| 12 | 5.01* (1.17) | 4.79* (1.23) | 2.87* (1.32) | 3.76* (1.30) | 2.64* (1.29) | **1.93** (1.39) |
| 13 | 4.88* (1.34) | 4.78* (1.35) | 3.03* (1.35) | 3.87* (1.21) | 2.53* (1.38) | **1.91** (1.14) |
| 14 | 5.06* (1.18) | 4.67* (1.23) | 3.22* (1.31) | 3.79* (1.33) | 2.45* (1.28) | **1.81** (1.25) |
| 15 | 5.01* (1.21) | 4.79* (1.32) | 2.83* (1.26) | 3.89* (1.09) | 2.54* (1.44) | **1.94** (1.23) |
| 16 | 4.84* (1.32) | 4.80* (1.28) | 2.88* (1.24) | 3.80* (1.46) | 2.58* (1.40) | **2.10** (1.29) |
| 17 | 4.76* (1.44) | 4.50* (1.32) | 3.05* (1.20) | 4.16* (1.32) | 2.47* (1.38) | **2.07** (1.49) |
| 18 | 4.84* (1.41) | 4.89* (1.02) | 3.14* (1.26) | 3.47* (1.45) | 2.68* (1.43) | **1.99** (1.39) |
| 19 | 5.07* (1.12) | 4.52* (1.38) | 3.25* (1.36) | 3.78* (1.36) | 2.38* (1.35) | **2.01** (1.31) |
| 20 | 4.93* (1.36) | 4.88* (1.00) | 3.13* (1.31) | 3.73* (1.30) | 2.48* (1.38) | **1.86** (1.19) |
| Average Rank | 4.97 | 4.76 | 3.05 | 3.76 | 2.51 | **1.95** |

Table A6. Comparison of six topical keyphrase extraction algorithms. Each value indicates the average rank by the participants for $T = 30$ in the Arxiv data set. The average rank of each algorithm is listed in the bottom row. Each value in a bracket indicates the standard deviation of the ranks. Each * tagged on the average rank represents statistically significant difference of the ranks between the corresponding algorithm and the proposed HSN, with the p-value being lower than 0.05.

| Topic | LDA | TF–ITF | N-gram TF–ITF | TR | TPR | **HSN (proposed)** |
|---|---|---|---|---|---|---|
| 1 | 5.02* (1.12) | 4.07* (1.28) | 3.10* (1.29) | 4.62* (1.27) | 2.31* (1.35) | **1.88** (1.19) |
| 2 | 4.52* (1.26) | 5.31* (1.02) | 3.53* (1.33) | 3.66* (1.26) | **1.79*** (1.00) | 2.19 (1.26) |
| 3 | 4.99* (1.17) | 4.93* (1.05) | 2.81* (1.59) | 3.51* (1.22) | 2.96* (1.29) | **1.81** (1.21) |
| 4 | 5.12* (1.06) | 4.87* (1.08) | 2.97* (1.29) | 3.67* (1.27) | 2.55* (1.40) | **1.82** (1.19) |
| 5 | 3.70* (0.99) | 5.05* (1.14) | 3.15* (1.24) | 5.05* (1.11) | 2.25* (1.26) | **1.81** (1.21) |
| 6 | 5.25* (1.07) | 3.94* (1.21) | 2.77* (1.15) | 4.71* (1.14) | 2.38* (1.35) | **1.94** (1.21) |
| 7 | 4.62* (1.17) | 4.29* (1.27) | 2.90* (1.10) | 5.15* (1.02) | 2.24* (1.22) | **1.77** (1.12) |
| 8 | 5.18* (1.15) | 2.84* (1.61) | 3.90* (1.21) | 4.66* (1.01) | 2.77* (1.12) | **1.66** (1.09) |
| 9 | 5.15* (1.04) | 3.69* (1.47) | 2.19* (1.28) | 4.61* (1.10) | 3.58* (1.15) | **1.77** (1.17) |
| 10 | 5.13* (1.03) | 4.78* (1.10) | 2.87* (1.10) | 3.70* (1.27) | 2.79* (1.66) | **1.74** (1.16) |
| 11 | 5.17* (0.97) | 4.79* (1.25) | 2.28* (1.38) | 3.81* (1.20) | 3.05* (1.24) | **1.90** (1.14) |
| 12 | 2.99* (1.11) | 5.07* (1.04) | 2.18* (1.23) | 5.18* (0.94) | 3.82* (1.18) | **1.77** (1.01) |
| 13 | 5.05* (1.19) | 4.81* (1.18) | 3.31* (1.21) | 3.70 (1.19) | 2.20* (1.36) | **1.92** (1.24) |
| 14 | 3.72* (1.16) | 4.89* (1.16) | 2.99* (1.16) | 5.14* (1.14) | 2.46* (1.32) | **1.80** (1.22) |
| 15 | 4.85* (1.05) | 3.82* (1.24) | 2.31* (1.39) | 5.15* (1.08) | 3.01* (1.25) | **1.86** (1.10) |
| 16 | 5.17* (1.03) | 4.73* (1.20) | 3.05* (1.22) | 4.01 (1.26) | 2.09* (1.07) | **1.95** (1.26) |
| 17 | 5.13* (1.17) | 3.91* (1.15) | 2.09* (1.03) | 5.03* (1.06) | 3.02* (1.12) | **1.82** (1.08) |
| 18 | 2.28* (1.34) | 5.18* (1.20) | 3.12* (1.21) | 4.64* (1.05) | 4.01* (1.24) | **1.77** (1.07) |
| 19 | 5.23* (1.04) | 3.79* (1.19) | 2.84* (1.23) | 4.78* (1.12) | 2.40* (1.35) | **1.96** (1.29) |
| 20 | 5.18* (0.99) | 3.97* (1.27) | 2.45* (1.49) | 4.67* (1.14) | 2.94* (1.18) | **1.78** (1.11) |
| 21 | 5.10* (1.20) | 4.03* (1.29) | 3.04* (1.11) | 4.70* (1.11) | 2.36* (1.39) | **1.77** (1.12) |
| 22 | 4.02* (1.29) | 4.71* (0.88) | 2.23 (1.40) | 5.18* (1.08) | 2.94* (1.27) | **1.90** (1.10) |
| 23 | 4.97* (1.34) | 3.75* (1.16) | 2.32 (1.43) | 4.86* (1.22) | 3.12* (1.19) | **1.97** (1.19) |
| 24 | 5.16* (1.07) | 4.17* (1.21) | 3.16* (1.18) | 4.56* (1.28) | 2.15* (1.26) | **1.80** (1.03) |
| 25 | 5.14* (1.16) | 4.66* (1.17) | 2.81* (1.21) | 3.97* (1.29) | 2.50* (1.37) | **1.92** (1.25) |
| 26 | 4.98* (1.02) | 3.69* (1.12) | 2.97* (1.21) | 5.13 (1.18) | 2.24* (1.30) | **1.99** (1.27) |
| 27 | 4.77* (1.01) | 3.79* (1.22) | 3.07* (1.25) | 5.17* (1.15) | 2.31* (1.40) | **1.88** (1.20) |
| 28 | 4.78* (1.11) | 3.94* (1.22) | 2.87* (1.17) | 5.15* (1.10) | 2.39* (1.41) | **1.87** (1.17) |
| 29 | 3.86* (1.21) | 2.00* (1.16) | 2.34* (1.42) | 5.20* (1.07) | 4.77* (1.17) | **1.94** (1.17) |
| 30 | 3.86* (1.10) | 4.95* (1.00) | 2.23* (1.31) | 5.24* (1.00) | 2.91* (1.07) | **1.81** (1.16) |
| Average Rank | 4.67 | 4.31 | 2.80 | 4.62 | 2.74 | **1.86** |

Figure B shows one of the survey questions that were provided to the participants. It consists of five topic-relevant research paper titles and the keyphrase sets obtained by the keyphrase extraction method. Each keyphrase set was produced from the same topical document collection.

Figure B. Survey example for $T$ = 10 in the Scopus dataset. Each participant was asked to rate the six keyphrase sets by LDA, TF–ITF, N-gram TF–ITF, TR, TPR, and the proposed HSN according to the five research paper titles.

**Q1.** Please rate the following keyword sets representing some research paper titles.
A keyword set is better than the other sets if the phrases (keywords) in the set include the meaning of all the titles.

| Research Paper Titles |
|---|
| Differences between adaptors and innovators in the context of entrepreneurial potential dimensions |
| The andragogical perspectives of older people's interaction with digital game technologies: Gameplay on gesture and touch based platforms |
| Identification of the learning behavior of students for education personalization |
| With a little help from my friends: a computational model for the role of social support in mood regulation |
| Lecturer perceptions of impoliteness and inappropriateness in student e-mail requests: a Norwegian perspective |

| ☐ | ☐ | ☐ | ☐ | ☐ | ☐ |
|---|---|---|---|---|---|
| 1 student | 1 social medium | 1 social | 1 student | 1 social network | 1 human perspective |
| 2 social | 2 social network | 2 model | 2 modelling | 2 social influence | 2 learning environment |
| 3 teacher | 3 learning environment | 3 system | 3 modeling | 3 social community | 3 social medium |
| 4 education | 4 college student | 4 paper | 4 modeler | 4 social study | 4 social networking |
| 5 school | 5 digital game | 5 datum | 5 social | 5 social medium | 5 social science |
| 6 educational | 6 mass medium | 6 user | 6 paper | 6 social dynamic | 6 collaborative learning |
| 7 game | 7 high school | 7 behavior | 7 researcher | 7 social behavior | 7 human behavior |
| 8 child | 8 entrepreneurial potential | 8 human | 8 researched | 8 social analysis | 8 social network |
| 9 cultural | 9 social interaction | 9 study | 9 study | 9 social interaction | 9 university student |
| 10 agent | 10 collaborative learning | 10 student | 10 educational | 10 social activity | 10 college student |